\documentclass[runningheads]{llncs}
\usepackage{graphicx}
\usepackage{comment}
\usepackage{amsmath,amssymb} 
\usepackage{color}
\usepackage{multirow}
\usepackage{epsfig}
\usepackage{caption}
\usepackage{hhline}
\usepackage{gensymb}
\usepackage{algorithm,algorithmicx,algpseudocode}
\usepackage{dsfont}
\usepackage{cite}
\usepackage[utf8]{inputenc}
\usepackage{upgreek}
\usepackage{hyperref}
\usepackage{verbatim}
\usepackage{nccmath}
\usepackage{comment}
\usepackage{color}
\usepackage{booktabs}
\usepackage{wrapfigure}
\usepackage{tikz}
\usepackage{comment} 
\usepackage{amsmath,amssymb} 
\usepackage{color}
\usepackage{float}

\newcommand{\todos}[1]{\textcolor{red}{{\em #1}}}
\usepackage[width=122mm,left=12mm,paperwidth=146mm,height=193mm,top=12mm,paperheight=217mm]{geometry}
\begin{document}
\pagestyle{headings}
\mainmatter
\def\ECCVSubNumber{1886}  

\title{Surface Normal Estimation of \textit{Tilted} Images\\via Spatial Rectifier} 


\titlerunning{Surface Normal Estimation of Tilted Images via Spatial Rectifier}
%
\author{Tien Do \and Khiem Vuong \and Stergios I. Roumeliotis \and Hyun Soo Park}
\index{Park, Hyun Soo}
\authorrunning{T. Do et al.}
%
\institute{University of Minnesota \\
\email{\{doxxx104|vuong067|stergios|hspark\}@umn.edu}
}
\maketitle

\begin{abstract}
In this paper, we present a spatial rectifier to estimate surface normals of tilted images. Tilted images are of particular interest as more visual data are captured by arbitrarily oriented sensors such as body-/robot-mounted cameras. Existing approaches exhibit bounded performance on predicting surface normals because they were trained using gravity-aligned images. Our two main hypotheses are: (1) visual scene layout is indicative of the gravity direction; and (2) not all surfaces are equally represented by a learned estimator due to the structured distribution of the training data, thus, there exists a transformation for each tilted image that is more responsive to the learned estimator than others. We design a spatial rectifier that is learned to transform the surface normal distribution of a tilted image to the rectified one that matches the gravity-aligned training data distribution. Along with the spatial rectifier, we propose a novel truncated angular loss that offers a stronger gradient at smaller angular errors and robustness to outliers. The resulting estimator outperforms the state-of-the-art methods including data augmentation baselines not only on ScanNet and NYUv2 but also on a new dataset called Tilt-RGBD that includes considerable roll and pitch camera motion.

\end{abstract}


\section{Introduction}
\begin{figure}[t]
    \begin{center}
    \includegraphics[width=\textwidth]{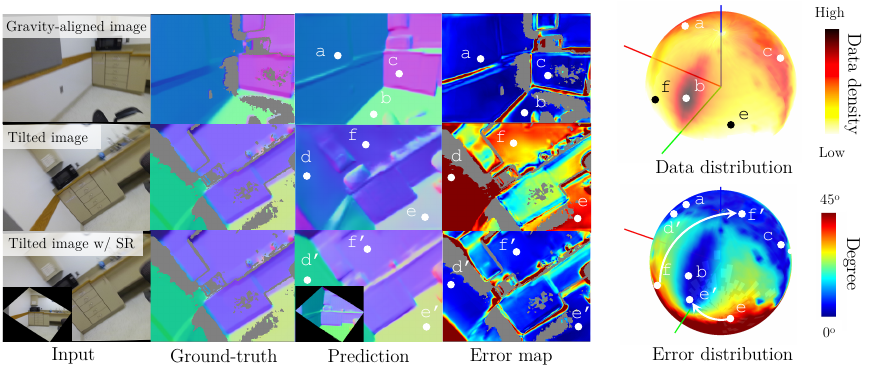}
    \end{center}
    \vspace{-2mm}
    \caption{We present a spatial rectifier that is designed to learn to warp a tilted image for surface normal estimation. While a state-of-the-art surface normal estimator~\cite{Huang19FrameNet} produces accurate prediction on an upright image (top row), it performs poorly on a tilted image (middle row) because of the bias in the surface normal distribution of the training data. Data distribution illustrates the density of surface normals in the ScanNet training dataset, which is highly correlated with the error distribution. We leverage the spatial rectifier to transform the surface normal in erroneous regions to nominal orientation regions, i.e., (\texttt{d},\texttt{e},\texttt{f})$\rightarrow$(\texttt{d'},\texttt{e'},\texttt{f'}) in error distribution, allowing accurate surface normal prediction (bottom row).
    }
    \label{fig:scannet_spherical_data_error_dist}
\end{figure}




We live within structured environments where various scene assumptions, e.g., Manhattan world assumption~\cite{NIPS2000_1804}, can be reasonably applied to model their 3D scene geometry. Nonetheless, the motion of daily-use cameras that observe these scenes is rather unrestricted, generating diverse visual data. For instance, \textit{embodied sensor measurements}, e.g., body-mounted cameras and agile robot-mounted cameras that are poised to enter our daily spaces, often capture images from tilted camera placement (i.e., non-zero roll and pitch). Understanding scene geometry, e.g., surface normal, from such non-upright imagery can benefit numerous real-world applications such as augmented reality~\cite{Huang19FrameNet, bansal2016Marr2D3DalignmentViaSN}, 3D reconstruction~\cite{Qi_2018_CVPR, zhang2018deepdepthcompletion, Liu2019planercnn, tang2019sparse2dense, weerasekera2017MonocularSN, qiu2019deeplidar}, and 3D robot navigation.

Existing learning-based frameworks, however, exhibit limited capability to predict surface normals of images taken from cameras with arbitrary pointing directions because not all scene surfaces are equally represented in deep neural networks.  Figure~\ref{fig:scannet_spherical_data_error_dist} shows the error distribution of surface normals predicted by a state-of-the-art network FrameNet~\cite{Huang19FrameNet} after being trained on the ScanNet dataset~\cite{dai2017scannet}. Albeit accurate on an upright image (regions \texttt{a}, \texttt{b}, and \texttt{c} in the top row), the prediction on a tilted image (regions \texttt{d}, \texttt{e}, and \texttt{f} in the middle row) is highly erroneous. This performance degradation is mainly caused by a  domain gap between the training and testing data, i.e., the training data is predominantly composed of gravity-aligned images while the testing tilted images have large roll and pitch angles. This poses a major limitation for surface normal estimation on embodied sensor measurements using the state-of-the-art estimator.

We address this challenge by leveraging a \textit{spatial rectifier}. Our main hypothesis is that it is possible to learn a global transformation that can transform a tilted image such that its surface normal distribution can be matched to that of gravity-aligned images. For instance, a homography warping induced by a pure 3D rotation can warp image features such that the surface normals in the underrepresented regions are transformed to the densely distributed regions (\texttt{d},\texttt{e},\texttt{f})$\rightarrow$(\texttt{d'},\texttt{e'},\texttt{f'}) as shown in Figure~\ref{fig:scannet_spherical_data_error_dist}. We design the spatial rectifier parameterized by the gravity direction and a principle direction of the tilted image that maximizes a distribution match, i.e., minimizes the domain gap between the tilted image and the upright training data. The spatial rectifier is learned by synthesizing tilted images using a family of homographies induced by the gravity and principle directions in an end-to-end fashion.
We demonstrate that the spatial rectifier is highly effective for surface normal estimation in embodied sensor data, outperforming the state-of-the-art methods including data augmentation baselines not only on ScanNet~\cite{dai2017scannet} and NYUv2~\cite{Silberman:ECCV12} but also a new dataset called Tilt-RGBD with a large roll and pitch camera motion. The bottom row of Figure~\ref{fig:scannet_spherical_data_error_dist} illustrates surface normal estimation with the spatial rectifier applied to a tilted image where the estimation error is substantially lower than the one without the spatial rectifier (the middle row).

Further, we integrate two new design factors to facilitate embodied sensor measurements in practice. (1)~Robustness: there exists a mismatch between the loss used in existing approaches and their evaluation metric. On the one hand, due to the noisy nature of ground truth surface normals~\cite{hickson2019floorsareflat}, the evaluation metric  relied on measures in the angle domain. On the other hand, $L_2$ loss or cosine similarity between 3D unit vectors is used to train the estimators for convergence and differentiability. This mismatch introduces sub-optimal performance of their normal estimators. We address this mismatch by introducing a new loss function called \textit{truncated angular loss} that is robust to outliers and accurate in the region of small angular error below 7.5$^\circ$. (2)~Resolution vs. context: Learning surface normal requires accessing  both global scene layout and fine-grained local features. Hierarchical representations, such as the feature pyramid network~\cite{Kirillov19PanopticFPN,fu2018dorn}, have been used to balance the representational expressibility and computational efficiency while their receptive fields are bounded by the coarsest feature resolution. In contrast, we propose a new design that further increases the receptive fields by incorporating dilated convolutions in the decoding layers. The resulting network architecture significantly reduces the computation (70\% FLOPS reduction as compared to the state-of-the-art DORN~\cite{fu2018dorn}) while maintaining its accuracy. This allows us to make use of a larger network (e.g., ResNeXt-101; 57\% FLOPS reduction) that achieves higher accuracy and faster inference (45 fps).

To the best of our knowledge, this is the first paper that addresses the challenges in surface normal estimation of tilted images without external sensors. Our contributions include: (1)~A novel spatial rectifier that achieves better performance by learning to warp a tilted image to match the surface normal distribution of the training data; (2)~A robust angular loss for handling outliers in the ground truth data; (3)~An efficient deep neural network design that accesses both global and local visual features, outperforming the state-of-the-art methods in computational efficiency and accuracy; and (4)~A new dataset called Tilt-RGBD that includes a large variation of roll and pitch camera angles measured by body-mounted cameras. 

\section{Related Work}

This paper includes three key innovations to enable accurate and efficient surface normal estimation for tilted images: Spatial rectification, robust loss, and network design. We briefly review most related work regarding datasets, accuracy, loss function, and network designs summarized in Table~\ref{table:related_works_table}.

\begin{table*}[t]
\centering
\footnotesize

\resizebox{\textwidth}{!}{\begin{tabular}{l|c|c|c|c|c|c}
\hline
Related Work     & Training data                                            & Testing data  & Loss & Backbone    & Median (\degree)~$\downarrow$ & 11.25\degree~$\uparrow$ \\ \hline\hline
Li et al. \cite{li2015depthnormalcrf}                   & \multirow{6}{*}{NYUv2}                                             & \multirow{8}{*}{NYUv2} & $L_2$          & AlexNet                          & 27.8                     & 19.6                    \\ \cline{1-1} \cline{4-7} 
Chen et al. \cite{Chen2017SurfaceNormalsInTheWilds}                  &                                                                     &                          & $L_2$          & \begin{tabular}[c]{@{}c@{}}AlexNet\end{tabular}   & 15.78                    & 39.17                   \\ \cline{1-1} \cline{4-7} 
Eigen and Fergus \cite{eigen2015DepthNormalSemanticLabel}                  &                                                                     &                          & $L_2$          & AlexNet/VGG                          & 13.2                     & 44.4                    \\ \cline{1-1} \cline{4-7} 
SURGE \cite{wang2016surge}                   &                                                                     &                          & $L_2$          & \begin{tabular}[c]{@{}c@{}}AlexNet/VGG$\times4$\end{tabular}           & 12.17                    & 47.29                   \\ \cline{1-1} \cline{4-7} 
Bansal et al. \cite{bansal2016Marr2D3DalignmentViaSN}                  &                                                                     &                          & $L_2$          & VGG-16                          & 12.0                     & 47.9                    \\ \cline{1-1} \cline{4-7} 
GeoNet \cite{Qi_2018_CVPR}                   &                                                                     &                          & $L_2$          & \begin{tabular}[c]{@{}c@{}}Deeplabv3\end{tabular} & 11.8                     & 48.4                    \\ \cline{1-2} \cline{4-7} 
FrameNet \cite{Huang19FrameNet} & \multirow{2}{*}{ScanNet}                                                            &                          & $L_2$          & \begin{tabular}[c]{@{}c@{}}DORN\end{tabular} & 11.0  & 50.7
\\ \cline{1-1} \cline{4-7} 
VPLNet \cite{wang2020vplnet} &                                                               &                          & AL          & \begin{tabular}[c]{@{}c@{}}UPerNet\end{tabular} & 9.83  & 54.3 \\ \hline
FrameNet \cite{Huang19FrameNet} & \multirow{3}{*}{ScanNet}                                            &     \multirow{3}{*}{ScanNet}                     & $L_2$          & \begin{tabular}[c]{@{}c@{}}DORN\end{tabular} & 7.7  & 62.5 \\ \cline{1-1} \cline{4-7} 
Zeng et al. \cite{zeng2019rgbdcompletion}                   &                                                                     &                          & $L_1$              & VGG-16                          & 7.47                  & 65.65                 \\ \cline{1-1} \cline{4-7} 
VPLNet \cite{wang2020vplnet}                   &                                                                     &                          & AL              & UPerNet                          & 6.68                  & 66.3                 \\ \hline
\end{tabular}}
\vspace{4mm}
\caption[Caption without FN]{Surface normal estimation methods and  performance comparison. 
The median measures the median of angular error and $11.25\degree$ denotes the percentage of pixels whose angular errors are less than $11.25\degree$ ($\downarrow/\uparrow$: the lower/higher the better). Note that Zeng et al. \cite{zeng2019rgbdcompletion} uses RGBD as an input. AL stands for angular loss (cos$^{-1}(\cdot)$).}\label{table:related_works_table}
\end{table*}

\noindent\textbf{Gravity Estimation and Spatial Rectification} Estimating gravity is one the fundamental problems in visual scene understanding. Visual cues such as vanishing points in indoor scene images~\cite{Mirzaei2011OptimalVPManhattan, Lee2015OrientationVPs} can be leveraged to estimate the gravity from images without external sensors such as an IMU. In addition, learning-based methods have employed visual semantics to predict gravity~\cite{fischer2015imageorientation, olmschenk2017pitchrollest, xian2019uprightnet}. This gravity estimate is, in turn, beneficial to recognize visual semantics, e.g., single view depth prediction~\cite{fei2019geosupervised, saito2020roll_depth}. In particular, a planar constraint enforced by the gravity direction and visual semantics (segmentation) is used to regularize depth prediction on KITTI~\cite{fei2019geosupervised}. Furthermore, a monocular SLAM method is used to correct in-plane rotation of images, allowing accurate depth prediction~\cite{saito2020roll_depth}. Unlike existing methods that rely on external sensors or offline gravity estimation, we present a spatial rectifier that can be trained in an end-to-end fashion in conjunction with the surface normal estimator. Our spatial rectifier, inspired by the spatial transformer networks~\cite{jaderberg2015stn}, transforms an image by a 3D rotation parameterized by the gravity and the principle directions.

\noindent\textbf{Robust Loss} There is a mismatch between the training loss and the evaluation metric for surface normal estimation. Specifically, while the cosine similarity or $L_2$ loss is used for training~\cite{bansal2016Marr2D3DalignmentViaSN, Chen2017SurfaceNormalsInTheWilds, eigen2015DepthNormalSemanticLabel, hickson2019floorsareflat, Huang19FrameNet, li2015depthnormalcrf, Qi_2018_CVPR, wang2016surge, zeng2019rgbdcompletion, zhan2019selfsupervisedLearningDepthSurfacenormal, zhang2018deepdepthcompletion}\footnote{Classification-based surface normal learning is not considered as its accuracy is bounded by the class resolution~\cite{Ladicky2014DiscriminativelySurfaceNormal, wang2015LocalGlobalSurfaceNormal}.}, the evaluation metric is, in contrast, based on the absolute angular error, e.g., $L_1$ or $L_0$. This mismatch causes not only slow convergence but also sub-optimality. Notably, Zeng et al.~\cite{zeng2019rgbdcompletion} and Liao et al.~\cite{liao2019spherical_regression} proposed to use a $L_1$ measure on unit vectors and spherical regression loss, respectively, to overcome the limitations of $L_2$ loss. More recently, UprightNet~\cite{xian2019uprightnet} and VPLNet~\cite{wang2020vplnet} employed an angular loss (AL),  and showed its high effectiveness in both gravity and surface normal predictions, respectively. In this work, we propose a new angular loss called the truncated angular loss that increases robustness to outliers in the training data.

\noindent\textbf{Efficient Network Design} Surface normal estimation requires to access both global and local features. Modern designs of surface normal estimators leverage a large capacity to achieve accurate predictions at high resolution. For instance, FrameNet~\cite{Huang19FrameNet} employs the DORN~\cite{fu2018dorn} architecture, a modification of DeepLabv3~\cite{chen2017deeplab} that removes multiple spatial reductions (2$\times$2 max pool layers) to achieve high resolution surface normal estimation at the cost of processing a larger number of activations. Recently, new hierarchical designs (e.g., panoptic feature pyramid networks~\cite{Kirillov19PanopticFPN}) have shown comparable performance by accessing both global and local features while significantly reducing computational cost in instance segmentation tasks. Our design is inspired by such hierarchical designs through atrous convolution in decoding layers, which outperforms DORN on acccuracy with lower memory footprint, achieving real-time performance.

\section{Method}
\label{sec:Method}
We present a new spatial rectifier to estimate the surface normals of tilted images. 
The spatial rectifier is jointly trained with the surface normal estimator by synthesizing randomly oriented images. 

\subsection{Spatial Rectifier}
\label{subsection:GAAT}
Given a tilted image $\mathcal{I}$, a surface normal estimator takes as input a pixel coordinate $\mathbf{x}\in \mathcal{R}(\mathcal{I})$ and outputs surface normal direction $\mathbf{n}\in\mathds{S}^2$, $f_{\boldsymbol{\phi}}(\mathbf{x};\mathcal{I}) = \mathbf{n}$, where $\mathcal{R}(\mathcal{I})$ is the coordinate range of the image $\mathcal{I}$. The surface normal estimator is parameterized by $\boldsymbol{\phi}$. 

Consider a spatial rectifier:
\begin{align}
    \overline{\mathcal{I}}(\mathbf{x}) = \mathcal{I}(W_{\mathbf{g},\mathbf{e}}(\mathbf{x})),
\end{align}
\begin{wrapfigure}{r}{0.22\textwidth}
  \centering
  \vspace{-10mm}
    \includegraphics[width=0.22\textwidth]{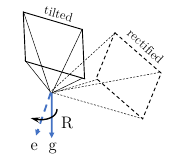}
  \caption{Geometry.} \label{Fig:geom}
\end{wrapfigure}
where $W_{\mathbf{g},\mathbf{e}}$ is the spatial rectifier that warps a tilted image $\mathcal{I}$ to the rectified image $\overline{\mathcal{I}}$, parameterized by the unit vectors $\mathbf{g}$ and $\mathbf{e}$ expressing the gravity and the principle directions, respectively, in the tilted image coordinate system (Figure~\ref{Fig:geom}). We define the spatial rectifier using a homography induced by a pure rotation mapping from $\mathbf{g}$ to $\mathbf{e}$, i.e., $W_{\mathbf{g},\mathbf{e}} = \mathbf{K}\mathbf{R}\mathbf{K}^{-1}$ where $\mathbf{K}$ is the matrix comprising the intrinsic parameters of the image $\mathcal{I}$. The rotation $\mathbf{R}$ can be written as:
\begin{align}
     \mathbf{R}(\mathbf{g},\mathbf{e}) = \mathbf{I}_{3} + 2\mathbf{e}\mathbf{g}^\mathsf{T} 
     - \frac{1}{1 +\mathbf{e}^\mathsf{T}\mathbf{g}} \left(\mathbf{e} + \mathbf{g}\right)\left(\mathbf{e}+\mathbf{g}\right)^\mathsf{T},\label{eq:rot_matrix_gravity_alignment} 
\end{align}
where $\mathbf{I}_3$ is the 3$\times$3 identity matrix. 

The key innovation of our spatial rectifier is a learnable $\mathbf{e}$ that transforms the surface normal distribution from an underrepresented region to a densely distributed region in the training data, i.e., (\texttt{d},\texttt{e},\texttt{f})$\rightarrow$(\texttt{d'},\texttt{e'},\texttt{f'}) in Figure~\ref{fig:scannet_spherical_data_error_dist}. In general, the principle direction determines the minimum amount of rotation of the tilted image towards the rectified one whose surface normals follow the distribution of the training data. A special case is when $\mathbf{e}$ is aligned with one of the image's principle directions, e.g., the y-axis or $\mathbf{e} = \begin{bmatrix}0 & 1& 0\end{bmatrix}^\mathsf{T}$, which produces a gravity-aligned image. We hypothesize that this principle direction and the gravity can be predicted by using the visual semantics of the tilted image:
\begin{align}
    h_{\boldsymbol{\theta}}(\mathcal{I}) = \begin{bmatrix} \mathbf{g}^\mathsf{T} & \mathbf{e}^\mathsf{T} \end{bmatrix} \label{eq:learnable_h},
\end{align}
where $h_{\boldsymbol{\theta}}$ is a learnable estimator parameterized by $\boldsymbol{\theta}$. 

\begin{figure}[t]
    \centering
    \includegraphics[width=1\textwidth]{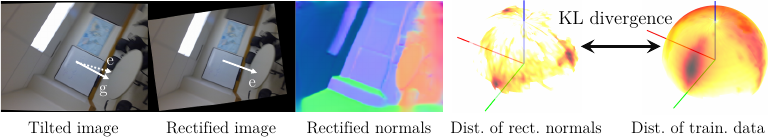}
    \caption{Spatial rectifier uses the tilted image to synthesize the rectified image by maximizing the surface normal distribution match. The rectified surface normal distribution is matched with that of the gravity aligned images.}
    \label{fig:rec1}
\end{figure}

We compute the ground truth of the principle direction $\mathbf{e}$ that maximizes the area of visible pixels in the rectified image while minimizing the KL divergence~\cite{Kullback:1951} of two surface normal distributions:
\begin{align}
    \mathbf{e}^* = \underset{\mathbf{e}}{\operatorname{argmin}}~D_{KL}\left(P(\mathbf{e})||Q\right) + \lambda_{\mathbf{e}}V\left(\overline{\mathcal{I}}\left(W_{\mathbf{g},\mathbf{e}}\right)\right), \label{Eq:e}
\end{align}
where $Q$ is the discretized surface normal distribution from all training data and $P(\mathbf{e})$ is the surface normal distribution of the rectified image given $\mathbf{e}$ (Figure~\ref{fig:rec1}). Minimizing the KL divergence allows us to find a transformation from the tilted image to the rectified image ($\mathbf{g}\rightarrow\mathbf{e}$), the surface normal distribution of which optimally matches the training data. $V$ counts the number of invisible pixels in the rectified image:
\begin{align}
    V(\overline{\mathcal{I}}) = \int_{\mathcal{R}(\overline{\mathcal{I}})} \delta \left(\overline{\mathcal{I}}(\mathbf{x})\right) {\rm d}\mathbf{x},
\end{align}
where $\delta(\cdot)$ is the Kronecker delta function that outputs one if the pixel at $\mathbf{x}$ is invisible, i.e., $\overline{\mathcal{I}}(\mathbf{x})=0$, and zero otherwise. This ensures a minimal transformation of the tilted image, facilitating learning surface normals from the rectified image. The scalar factor $\lambda_{\mathbf{e}}$ balances the relative importance of distribution matching and visibility. Figure~\ref{fig:rec1} illustrates the optimal $\mathbf{e}$ that maximizes the surface normal distribution match between the rectified image and the gravity-aligned training images (e.g., ScanNet~\cite{dai2017scannet}). 

\subsection{Surface Normal Estimation by Synthesis}
\label{subsection:surface_normal_synthesis}
\begin{figure}[t]
    \centering
    \includegraphics[width=1\textwidth]{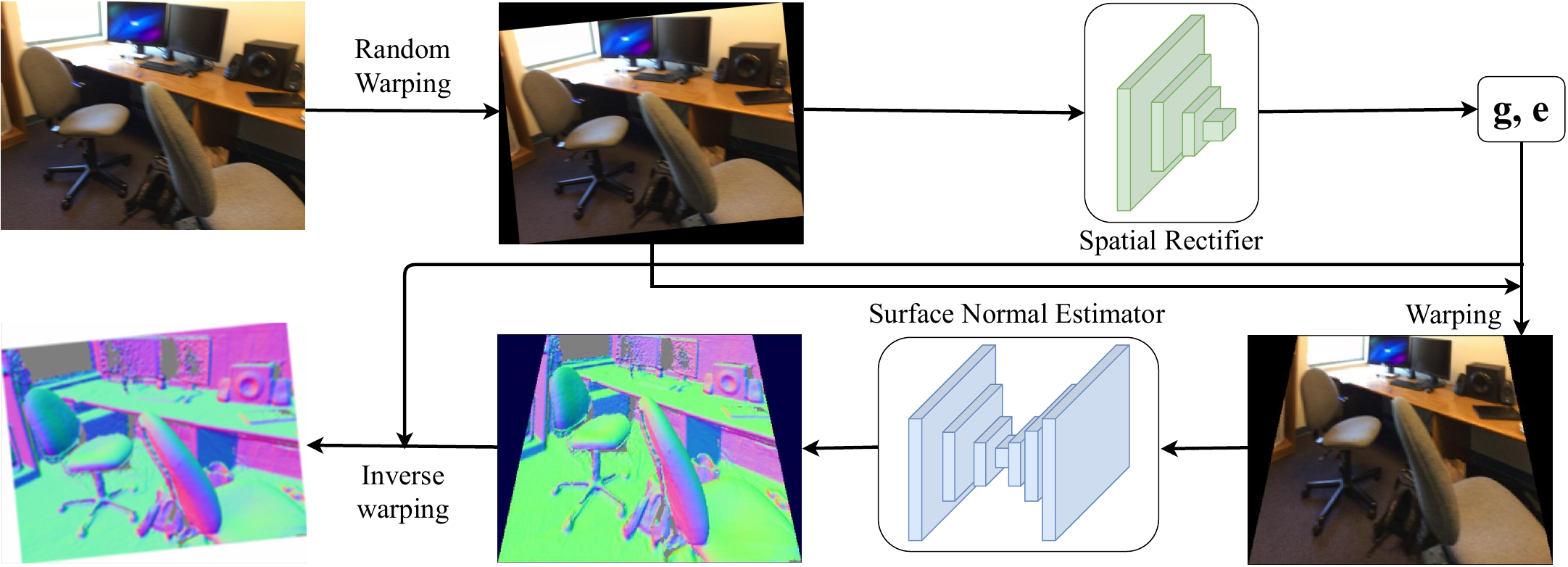}
    \caption{We jointly train the spatial rectifier and surface normal estimator by randomly synthesizing tilted images. A training image is warped to produce a tilted image. The rectifier is learned to predict $\mathbf{g}$ and $\mathbf{e}$ in the tilted image coordinate system. This allows us to warp the tilted image to the rectified image whose surface normal distribution matches that of the training data. The surface normal estimator is used to predict the rectified surface normals and warp back to the tilted  surface normals.}
    \label{fig:full}
\end{figure}

We learn the parameters of the neural networks ($\boldsymbol{\phi}$ and $\boldsymbol{\theta}$) by minimizing the following loss:
%
\begin{align}
    \mathcal{L}(\boldsymbol{\phi}, \boldsymbol{\theta}) = \sum_\mathcal{D} \sum_{\mathbf{x}\in \mathcal{R}(\mathcal{I})} \mathcal{L}_{\rm TAL}(\mathbf{n}_{\mathbf{x};\mathcal{I}}, \widehat{\mathbf{n}}_{\mathbf{x};\mathcal{I}}) + \lambda \mathcal{L}_{\rm TAL}(h_{\boldsymbol{\theta}}(\mathcal{I}), \begin{bmatrix} \widehat{\mathbf{g}}^\mathsf{T} & \widehat{\mathbf{e}}^\mathsf{T} \end{bmatrix}), 
\end{align}
where $\mathcal{D}$ is the training dataset. For each training image, we synthesize a tilted image $\mathcal{I}$ and its corresponding ground truth surface normals $\widehat{\mathbf{n}}_{\mathbf{x};\mathcal{I}}$,  gravity $\widehat{\mathbf{g}}$, and principle axis $\widehat{\mathbf{e}}$. $\lambda$ controls the importance between the surface normal and the spatial rectification. $\mathcal{L}_{\rm TAL}$ is the truncated angular loss (Section~\ref{Sec:loss}). 
%
The predicted surface normals of the tilted image $\mathbf{n}_{\mathbf{x};\mathcal{I}}$ is obtained by transforming the rectified one $f_{\boldsymbol{\phi}}(\mathbf{x};\overline{\mathcal{I}})$ as follows: 
\begin{align*}
\mathbf{n}_{\mathbf{x};\mathcal{I}}= \mathbf{R}^{\mathsf{T}}_{\mathbf{g}, \mathbf{e}} f_{\boldsymbol{\phi}}(\mathbf{W}^{\mathsf{ -1}}_{\mathbf{g}, \mathbf{e}}(\mathbf{x});\overline{\mathcal{I}}).
\end{align*}
Note that since $f_{\boldsymbol{\phi}}$ takes rectified images as inputs, it does not require a large capacity network to memorize all possible tilted orientations.
Figure~\ref{fig:full} illustrates our approach that  synthesizes a tilted image with a random rotation where its predicted parameters ($\mathbf{g}$ and $\mathbf{e}$) are used to rectify the synthesized image. We measure the loss of surface normals by transforming back to the tilted image coordinate.

\subsection{Truncated Angular Loss} \label{Sec:loss}
$L_2$ loss on 3D unit vectors was used to train a surface normal estimator in the literature~\cite{bansal2016Marr2D3DalignmentViaSN, Chen2017SurfaceNormalsInTheWilds, eigen2015DepthNormalSemanticLabel, hickson2019floorsareflat, Huang19FrameNet, li2015depthnormalcrf, Qi_2018_CVPR, wang2016surge, zeng2019rgbdcompletion, zhan2019selfsupervisedLearningDepthSurfacenormal, zhang2018deepdepthcompletion}:

\begin{align*}
 \mathcal{L}_{L_2}(\mathbf{n}, \widehat{\mathbf{n}}) &= \frac{1}{2}\| \mathbf{n}- \widehat{\mathbf{n}}\|_{2}^{2} = 1 - \mathbf{n}^{\mathsf{T}}\widehat{\mathbf{n}} \label{eq:traditional_supervised_loss}
\end{align*}
where $\mathbf{n}$ and $\widehat{\mathbf{n}}$ are the estimated surface normals and its ground truth, respectively. This measures Euclidean distance between two unit vectors, which is equivalent to the cosine similarity. A main limitation of this loss is the vanishing gradient around small angular error, i.e., $\partial \mathcal{L}_{L_2} / \partial \xi \approx 0$ if $\xi \approx 0$, where $\xi=\text{cos}^{-1}(\mathbf{n}^{\mathsf{T}}\widehat{\mathbf{n}})$. As a results, the estimator is less sensitive to small angular error.

On the other hand, the angular error or the cardinality of the pixel set with an angular threshold has been used for their evaluation metric to address considerable noise in the ground truth surface normals~\cite{hickson2019floorsareflat}, i.e.,
\begin{align*}
    E_{\rm angular} &= \cos^{-1}(\mathbf{n}^{\mathsf{T}}\widehat{\mathbf{n}}), ~~~{\rm or}\\
    E_{\rm card}(\xi) 
    &= |\{\cos^{-1}(\mathbf{n}^{\mathsf{T}}\widehat{\mathbf{n}}) < \xi \}|, \ \ \xi = 11.25^{\circ}, 22.5^{\circ}, 30^{\circ} 
\end{align*}

This mismatch between the evaluation metric and the training loss introduces sub-optimality in the trained estimator. To address this issue, we propose the truncated angular loss (TAL) derived from the angular loss:
%
\begin{align}
    \mathcal{L}_{\rm TAL}(\mathbf{n},\widehat{\mathbf{n}}) = 
 \begin{cases}
    0, \ \ \ \ \ \ \ \ \ \ \ \ \ \ \ 1-\epsilon \leq & \mathbf{n}^\mathsf{T}\widehat{\mathbf{n}} \\
    \cos^{-1}(\mathbf{n}^\mathsf{T}\widehat{\mathbf{n}}), \ \ \ \ \ \ 0 \leq & \mathbf{n}^\mathsf{T}\widehat{\mathbf{n}} < 1-\epsilon \\
    \frac{\pi}{2} - \mathbf{n}^\mathsf{T}\widehat{\mathbf{n}} , & \mathbf{n}^\mathsf{T}\widehat{\mathbf{n}} < 0
  \end{cases} 
 \end{align}
\begin{wrapfigure}{l}{0.4\textwidth}
\vspace{-9mm}
  \centering
    \includegraphics[width=0.4\textwidth]{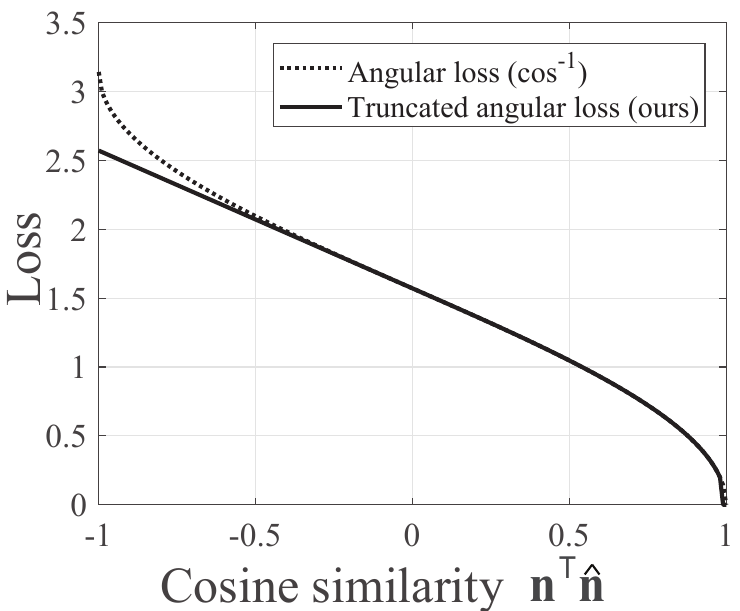}
  \caption{Truncated angular loss.} \label{Fig:loss}
  \vspace{-5mm}
\end{wrapfigure}
\noindent where $\epsilon=10^{-6}$. When $\mathbf{n}^\mathsf{T}\widehat{\mathbf{n}}\approx 1$, i.e., as the angular error is close to zero, the loss is clamped zero to avoid the infinite gradient of inverse cosine. More importantly, when $\mathbf{n}^\mathsf{T}\widehat{\mathbf{n}} < 0$, i.e., where the angular error is large (outlier), TAL is linear, i.e., it assigns the constant weight for each  $\mathbf{n}^\mathsf{T}\widehat{\mathbf{n}}$ in this region, similar to $L_2$, whereas AL assigns larger weight for larger $\mathbf{n}^\mathsf{T}\widehat{\mathbf{n}}$, as shown in Figure~\ref{Fig:loss}. This makes the training with TAL less sensitive to outliers than AL. \\
\subsection{Surface Normal Estimator Design}
\label{subsection:network_architecture}

For the surface normal estimator module, we design a new network architecture inspired by the Panoptic FPN~\cite{Kirillov19PanopticFPN} that has shown strong performance on high-resolution prediction tasks while maintaining its compact representation. We employ an asymmetric encoder-decoder structure with the following modifications: (i)~we do not suppress the number of channels of each pyramid feature immediately to $C=128$; and (ii)~we make use of the atrous spatial pyramid pooling (ASPP) module similar to DeepLabv3~\cite{chen2017deeplab}. 


%

\begin{figure}[t]
    \centering
    \includegraphics[width=1\textwidth]{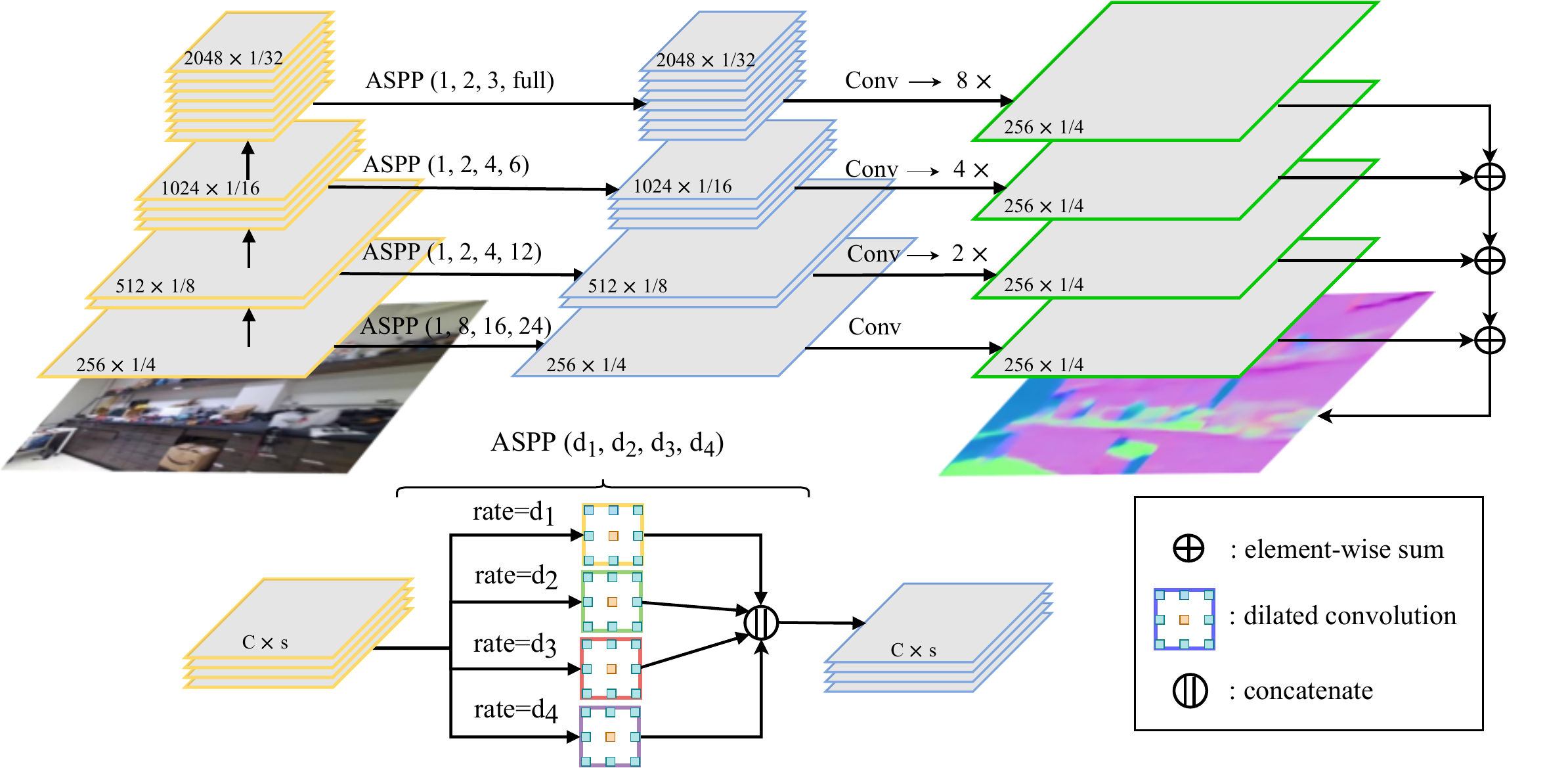}
    \caption{Illustration of the surface normal estimation network architecture.}
    \label{fig:network_architecture}
\end{figure}

The network encodes an image with multi-scale feature maps ranging from $1/4$ to $1/32$ for spatial resolution and $256$ to $2048$ for channel dimension. For the decoder, we assemble these multi-scale features with ASPP to increase the receptive fields as shown in Figure~\ref{fig:network_architecture}. The decoded feature maps are combined and upsampled to produce a quarter resolution feature maps and summed to predict surface normals.

\section{Results}

We evaluate the performance of our surface normal estimation quantitatively and qualitatively. All networks used in the evaluation including ours take as input $320\times240$ resolution image and output the same size surface normal. $\mathbf{h}_{\boldsymbol{\theta}}$ makes use of ResNet-18 backbone. The networks are trained with a batch size of 16 and optimized by the Adam~\cite{kingma1412adam} optimizer with a learning rate $10^{-4}$. We use an NVIDIA Tesla V100 GPU (with 32GB of memory) to train for 20 epochs and report the best epoch on the corresponding dataset’s validation set. Our code, dataset and pretrained networks are available at \url{https://github.com/MARSLab-UMN/TiltedImageSurfaceNormal}.


\noindent \textbf{Evaluation Metrics} For the quantitative evaluation of surface normal prediction, we employ standard metrics used in~\cite{bansal2016Marr2D3DalignmentViaSN, Fouhey_2013_ICCV}: (a) mean absolute of the error (mean), (b) median of absolute error (median), (c) root mean square error (RMSE), and (d) the percentage of pixels with angular error below a threshold $\xi$ with $\xi = $ 5\degree, 7.5\degree, 11.25\degree, 22.5\degree, 30.0\degree.

\begin{figure}[t]
    \includegraphics[width=1\textwidth]{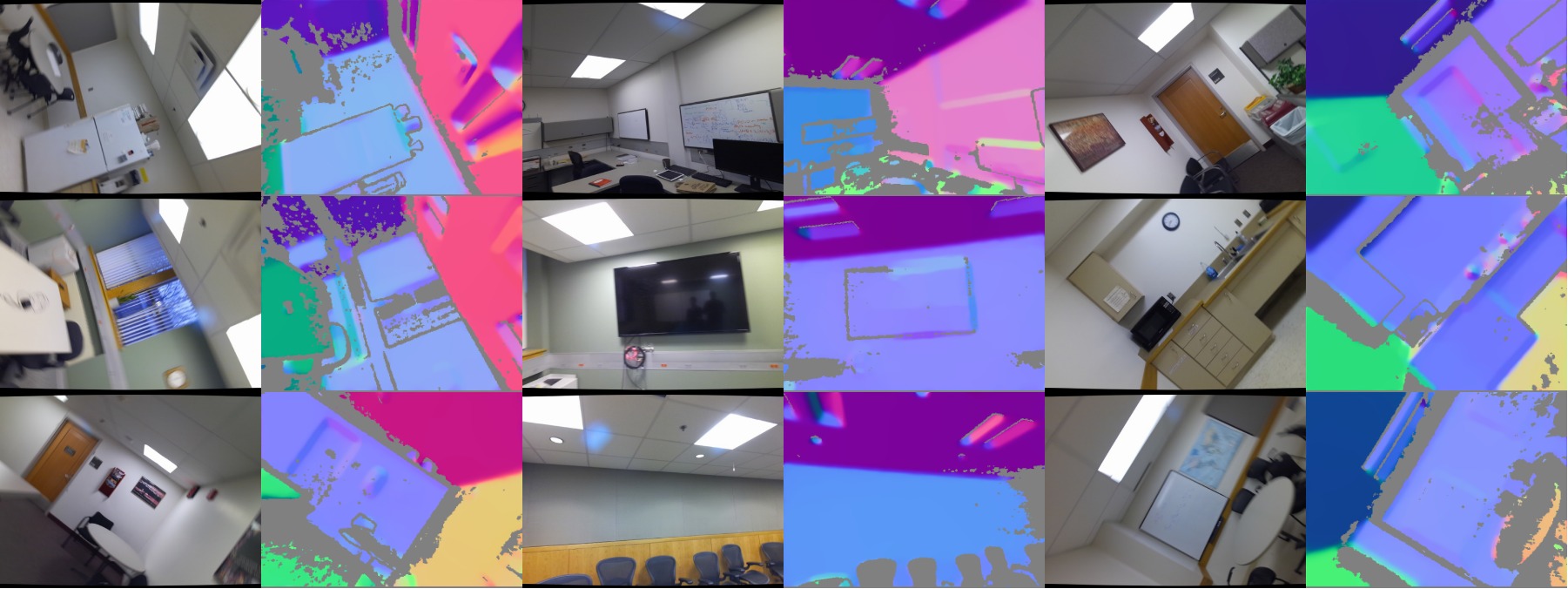}
    \caption{Selected RGB images and ground truth surface normals from the Tilt-RGBD dataset.}
    \label{fig:tilt-rgbd}
\end{figure}
\subsection{Evaluation Dataset}
We evaluate our method using ScanNet~\cite{dai2017scannet}, NYUv2~\cite{Silberman:ECCV12}, and a new dataset from body-mounted cameras called \textit{Tilt-RGBD} that includes tilted images. All networks are trained and validated by ScanNet and tested on NYUv2 and Tilt-RGBD. No domain adaptation is applied.

\noindent \textbf{ScanNet}  ScanNet~\cite{dai2017scannet} is an indoor RGB-D video dataset that spans a large variation of scenes. We use the images from their standard training/validation/testing splits by extracting every 10$^{\rm th}$ frame: 189,916 images for training, 53,193 images for validation, and 20,942 images for testing. For the ground truth surface normals, we make use of the ground truth generated by FrameNet~\cite{Huang19FrameNet}. For comparison with FrameNet, we make use of the training (199,720)/testing (64,319) split that was originally used in the paper.

\noindent \textbf{NYUv2} NYUv2~\cite{Silberman:ECCV12} was captured by MS Kinect, which includes 654 testing images. The ground truth surface normals are generated by Ladicky et al. \cite{Ladicky2014DiscriminativelySurfaceNormal}. Note that we evaluate only on the valid pixels following Zhang et al. \cite{zhang2017physically}.

\noindent \textbf{Tilt-RGBD} We collect a new dataset with body-mounted cameras (Azure Kinect) that includes 24 different scenes. Each image instance is associated with the gravity measured by an IMU following~\cite{do2019gyroless}. Two types of scenes are collected. (i)~\emph{Gravity-aligned images}: We control the camera orientation to be aligned with gravity. 14 scenes and 2,403 images are used. (ii)~\emph{Tilted images}: We capture images without controlling the camera's pose; this process generates a large variation of tilted images that includes 14 scenes and 2,428 images. We follow \cite{Ladicky2014DiscriminativelySurfaceNormal} approach to generate ground truth surface normals from the depth images. Figure~\ref{fig:tilt-rgbd} shows  representative RGB images along with their ground truth surface normals.

\subsection{Baseline}
\noindent \textbf{FrameNet} FrameNet~\cite{Huang19FrameNet} uses the DORN design that jointly predicts surface normal, tangent and bitangent directions.
We obtain the provided pre-trained network on ScanNet to perform our experiments;
\noindent \textbf{VPLNet} VPLNet~\cite{wang2020vplnet} incorporates lines and vanishing point directions to improve surface normals prediction, and is currently the state-of-the-art surface normal estimator on ScanNet and NYUv2.
Note that we are only able to compare with VPLNet on ScanNet and NYUv2 using the reported results~\cite{wang2020vplnet} since neither the pretrained network nor the implementation is provided.
\textbf{DORN+TAL} DORN~\cite{fu2018dorn} is a high-capacity network based on the ResNet-101~\cite{He_2016_CVPR} backbone. We use our truncated angular loss (TAL) to train the network; \textbf{DORN+TAL+SR} We train DORN with our proposed spatial rectifier (SR) described in Section~\ref{subsection:surface_normal_synthesis}; \textbf{DFPN+TAL} Dilated feature pyramid network (DFPN) is our proposed network described in Section~\ref{subsection:network_architecture} with our truncated angular loss (TAL). We use DFPN with ResNet-101 backbone; \textbf{DFPN+TAL+AUG} We train our DFPN with data augmentation (AUG) by synthesizing tilted images with random rotations. This network does not take rectified images and therefore, is expected to memorize all the tilted representations; \textbf{DFPN+TAL+IMU} We train our DFPN with rectified images by assuming the gravity is given (from the ground plane normal direction for ScanNet);  \textbf{DFPN+TAL+SR (ours)} We train our DFPN with our spatial rectifier (SR) described in Section~\ref{subsection:surface_normal_synthesis}.



\subsection{Surface Normal Estimation on Tilt-RGBD}
\label{subsection:res_generalization}

\begin{table}[t]
\setlength\aboverulesep{0pt}
\setlength\belowrulesep{0pt}
\centering
{\scriptsize
\scalebox{1.24}{
\begin{tabular}{@{}l|cccccccc@{}}
\toprule
Gravity-aligned images & Mean  & Median & RMSE  & 5\degree  & 7.5\degree & 11.25\degree & 22.5\degree & 30.0\degree \\ \midrule
FrameNet                 & 11.57 & 6.59   & 18.19 & 37.68 & 55.69  & 71.23    & 86.85   & 90.93   \\
DORN+TAL                     & 9.54  & 4.96   & 16.60 & 50.33 & 67.56  & 79.74    & 90.24   & 92.96   \\
DORN+TAL+SR                     & 9.40  & 5.49   & 15.76 & 45.09 & 65.75  & 80.71    & 91.31   & 93.75   \\
DFPN+TAL                     & 9.74  & 5.40   & 16.30 & 46.05 & 65.45  & 78.90    & 90.40   & 93.14   \\
DFPN+TAL+AUG & 10.19 & 5.62   & 17.57 & 43.89 & 64.39  & 79.07    & 89.95   & 92.62   \\
DFPN+TAL+IMU & 8.62  & \textbf{4.26}   & 15.57 & \textbf{56.90} & \textbf{72.42}  & 82.03    & 91.17   & 93.68   \\
DFPN+TAL+SR       & \textbf{8.59}  & 4.42   & \textbf{15.24} & 55.80 & 72.37  & \textbf{82.26}    & \textbf{91.39}   & \textbf{94.01}   \\ 
\toprule
Tilted images & Mean  & Median & RMSE  & 5\degree  & 7.5\degree & 11.25\degree & 22.5\degree & 30.0\degree \\ \midrule
FrameNet                 & 17.65 & 10.51  & 25.38 & 24.62 & 38.35  & 52.25    & 73.65   & 81.17   \\
DORN+TAL                     & 14.33 & 7.09   & 22.76 & 37.15 & 52.08  & 65.03    & 80.92   & 85.79   \\
DORN+TAL+SR                     & 11.77 & 6.59   & 18.95 & 37.00 & 56.05  & 72.35    & 87.29   & 90.72   \\
DFPN+TAL                    & 14.53 & 7.84   & 22.47 & 32.01 & 48.30  & 62.92    & 81.41   & 86.43   \\
DFPN+TAL+AUG & 12.89 & 6.86   & 21.28 & 34.81 & 54.28  & 71.20    & 85.75   & 89.10   \\
DFPN+TAL+IMU & 11.46 & \textbf{5.22}   & 19.88 & \textbf{48.30} & \textbf{62.92}  & 74.24    & 86.29   & 89.75   \\
DFPN+TAL+SR       & \textbf{11.22} & 5.82   & \textbf{18.88} & 43.03 & 61.19  & \textbf{75.02}    & \textbf{87.53}   & \textbf{90.80}   \\
\bottomrule
\end{tabular}}
}
\vspace{4mm}
\caption{Comparison between all baselines in term of generalization capacity on Tilt-RGBD on gravity-aligned and tilted images.}
\label{table:generalization_performance_}
\end{table}

We compare our method with the baselines on Tilt-RGBD on both gravity-aligned and tilted images summarized in  Table~\ref{table:generalization_performance_}. We observe that all networks trained on ScanNet perform excellent on unseen gravity-aligned frames indicating that the ScanNet dataset contains sufficient scene diversity. Nevertheless, the best performance in terms of median and tight thresholds ($\xi = 5\degree, 7.5\degree$) belongs to DFPN+TAL+IMU. This is due to the fact that gravity direction estimated from the IMU is highly accurate and is employed directly as the surface normal direction of the dominating part of the scene, e.g., floors, ceilings. Our method achieves on-par performance with DFPN+TAL+IMU since the gravity estimate from the network is less accurate.

\begin{figure}[t]
    \centering
    \includegraphics[width=\textwidth]{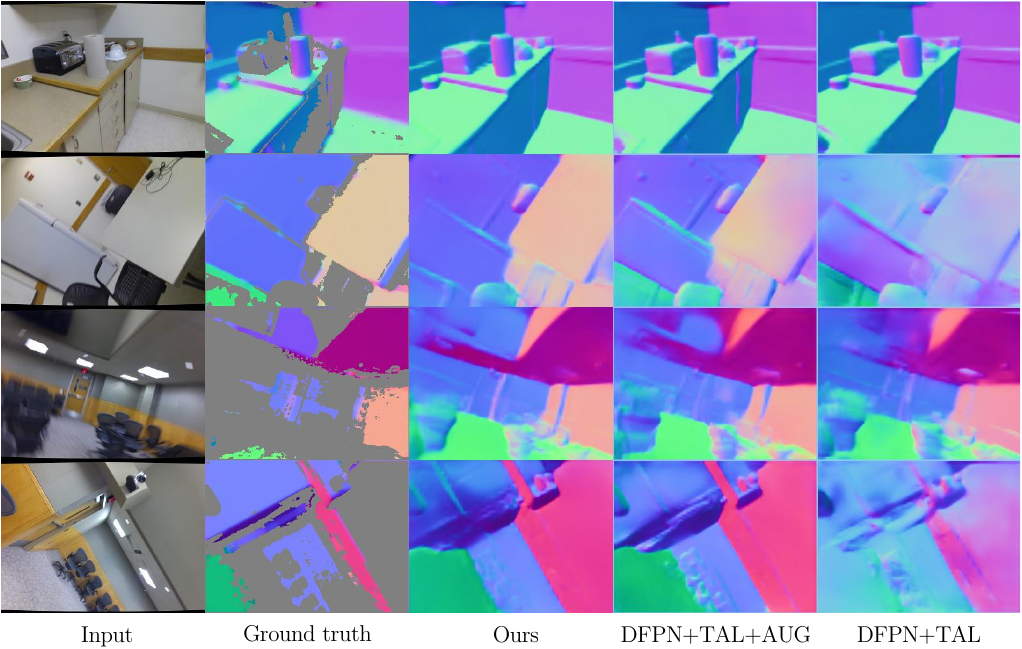}
    \caption{Qualitative results: We compare ours (DFPN+TAL+SR) with DFPN+TAL+AUG and DFPN+TAL.}
    \label{fig:qualitative}
\end{figure}

For tilted images, there is a significant performance degradation in all metrics, e.g., the percentage drop in $11.25\degree$ for FrameNet ($\sim$19\%), DFPN+TAL ($\sim$16\%), DORN+TAL ($\sim$14\%) while DORN+TAL+SR, DFPN+TAL+AUG, DFPN+TAL+IMU, and DFPN+TAL+SR show less degradation due to either data augmentation or external sensor information. DFPN+TAL+SR performs significantly better than augmenting the data DFPN+TAL+AUG and on par with DFPN+TAL+IMU while it does not require an external sensor. In addition, DFPN+TAL+SR also outperforms DORN+TAL+SR although DORN has higher capacity than our DFPN (Section~\ref{subsection:net_architecture}), which suggests that the SR is more compatible with our DFPN. Figure~\ref{fig:qualitative} shows our qualitative results where we compare our method with the baselines.

\subsection{Network Efficiency}
\label{subsection:net_architecture}
We compare our proposed DFPN+TAL with DORN+TAL
in terms of number of floating operations (FLOPS), number of parameters, actual memory consumption, and inference time summarized in Table~\ref{table:network_flops_accuracy} as well as surface normal 
\begin{wraptable}{r}{0.65\textwidth}
\vspace{-7mm}
\setlength\aboverulesep{0pt}
\setlength\belowrulesep{0pt}
\scriptsize
\scalebox{1.05}{
\begin{tabular}{l|l|cccc}
\toprule
Network  & Backbone & \# params & FLOPS & Memory & FPS \\ \hline
P-FPN+TAL & ResNet-101  & 53.5M        & 17.3G  & 6.9GB  &  68.5 \\
DORN+TAL     & ResNet-101  & 99.5M        & 97.0G  & 13.2GB  &  18.6 \\
DFPN+TAL     & ResNet-101  & 93.0M        & 29.2G  & 7.8GB   &  43.7  \\
DFPN+TAL     & ResNeXt-101 & 137.3M       & 42.6G  & 11.5GB  &  30.0 \\ \bottomrule
\end{tabular}
}
\caption{Comparison between different network architectures in terms of FLOPS, number of parameters, memory consumption (in training), and inference time (FPS with batch size of 4).}
\label{table:network_flops_accuracy}
\vspace{-5mm}
\end{wraptable}
estimation accuracy in Table~\ref{table:network_statistics}. With the ResNet-101 backbone, our DFPN +TAL shows significant improvement in terms of accuracy over the Panoptic FPN (P-FPN)~\cite{Kirillov19PanopticFPN} and performs comparably to DORN+TAL while highly efficient (less than 1/3 of DORN+TAL in terms of FLOPs, memory, and inference time (FPS)). With efficient DFPN+TAL, we further increase the network capacity to ResNeXt-101~\cite{xie2017aggregated}, producing more accurate prediction. It outperforms DORN+TAL in terms of accuracy both on ScanNet and NYUv2 with smaller FLOPS (2.4$\times$ faster training time), memory consumption, and realtime inference (30 FPS on NVIDIA GTX 1660). 
\begin{table}[ht]
\setlength\aboverulesep{0pt}
\setlength\belowrulesep{0pt}
\small
\scalebox{0.8}{
\begin{tabular}{@{}l|l|cccccc|cccccc@{}}
\toprule
\multirow{2}{*}{Network} & \multirow{2}{*}{Backbone} & \multicolumn{6}{c|}{\textbf{ScanNet}}                                                        & \multicolumn{6}{c}{\textbf{NYUv2}}                                                          \\ \cmidrule(){3-14} 
                         &                           & Mean          & Median       & RMSE          & 11.25\degree      & 22.5\degree       & 30\degree         & Mean          & Median       & RMSE          & 11.25\degree      & 22.5\degree       & 30\degree         \\ \midrule
P-FPN+TAL                 & ResNet-101                & 15.8         & 8.0         & 24.8         & 61.8         & 78.8         & 84.2         & 17.0         & 8.9         & 26.0         & 57.2         & 75.6         & 82.1         \\
DORN+TAL                     & ResNet-101                & 15.1          & \textbf{7.4} & 24.1          & \textbf{63.8} & 79.6          & 84.9          & 16.6          & 8.5          & 25.7          & 58.5          & 76.4          & 82.6          \\
DFPN+TAL                     & ResNet-101                & 15.4          & 7.5          & 24.6          & 63.3          & 79.4          & 84.7          & 16.9          & 8.6          & 26.2          & 58.2          & 76.1          & 82.2          \\
DFPN+TAL                     & ResNeXt-101               & \textbf{15.0} & 7.5          & \textbf{23.9} & \textbf{63.8} & \textbf{80.0} & \textbf{85.2} & \textbf{16.2} & \textbf{8.2} & \textbf{25.3} & \textbf{59.5} & \textbf{77.1} & \textbf{83.2} \\ \bottomrule
\end{tabular}}
\vspace{4mm}
\caption{Accuracy comparison between different network architectures.}
\label{table:network_statistics}
\end{table}
\subsection{Surface normal training loss}
\label{subsection:training_loss}
\begin{figure}[t]
    \centering
    \includegraphics[width=\textwidth]{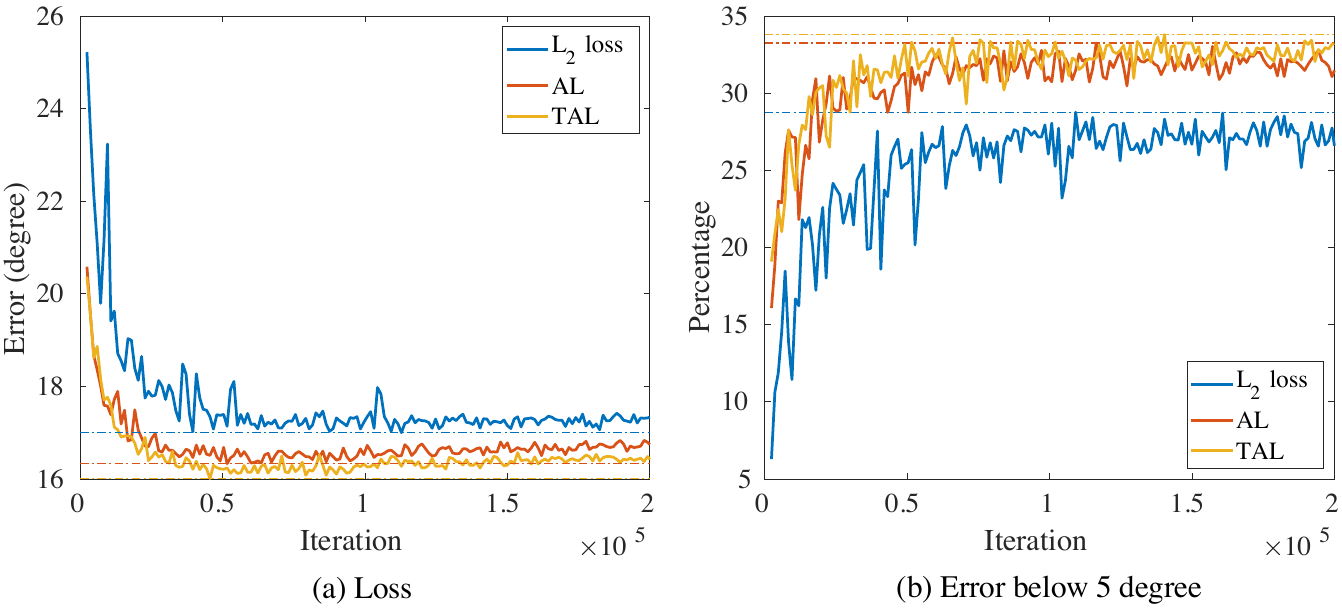}
    \caption{(a) Loss values and (b) percentage of pixels with angular error less than 5$^\circ$ over training iterations evaluated on NYUv2. Truncated angular loss (TAL) outperforms $L_2$ and Angular loss (AL).}
    \label{fig:rec}
\end{figure}

%
We compare three losses used for surface normal estimation: $L_2$, angular loss (AL), and truncated angular loss (TAL). 
%
%
As shown in Figure~\ref{fig:rec} and Table~\ref{table:loss_statistics}~(Top), changing from $L_2$ to AL or TAL leads to faster convergence and significant improvement in all of the error metrics. 
%
%
Furthermore, the TAL loss performs on-par with AL at loose thresholds such as 22.5\degree and 30\degree and shows improvement in particular at tight thresholds such as 5\degree and 7.5\degree, both on ScanNet and NYUv2.
In addition, we also show in the Table~\ref{table:loss_statistics}~(Bottom) that our method outperforms FrameNet and VPLNet in all of the metrics both on ScanNet and NYUv2. 
Note that in order to ensure the fairness in comparison with FrameNet~\cite{Huang19FrameNet} and VPLNet~\cite{wang2020vplnet}, we train and evaluate our method on their modified ScanNet data split.
Our results suggest that TAL is highly beneficial than the $L_2$ and AL losses, which allows us to achieve the state-of-the-art performance across different train/test splits and also maintains its generalization capability. 
%

\begin{table}[t]
\setlength\aboverulesep{0pt}
\setlength\belowrulesep{0pt}
\scalebox{0.74}{
\begin{tabular}{@{}l|cccccccc|cccccccc@{}}
\toprule
\multirow{2}{*}{Loss/Method } & \multicolumn{8}{c|}{\textbf{ScanNet}}                                                                                        & \multicolumn{8}{c}{\textbf{NYUv2}}                                                                                          \\ \cmidrule(){2-17} 
                              & Mean          & Median       & RMSE          & 5\degree          & 7.5\degree        & 11.25\degree      & 22.5\degree       & 30\degree         & Mean          & Median       & RMSE          & 5\degree          & 7.5\degree        & 11.25\degree      & 22.5\degree       & 30\degree         \\ \midrule
$L_2$                             & 15.7  & 8.2    & 24.3  & 30.4     & 46.7    & 61.2    & 78.7 & 84.4 & 17.1 & 9.4 & 25.7 & 28.0 & 42.1 & 55.7 & 75.2 & 82.0          \\
AL                           & 15.1  & 7.5    & 24.0  & 33.9     & 50.1    & 63.8    & 79.9 & 85.2 & 16.7 & 8.6 & 25.6 & 30.9 & 45.3 & 58.3 & 76.6 & 82.8          \\
TAL                    & \textbf{15.0} & \textbf{7.5} & \textbf{23.9} & \textbf{34.2} & \textbf{50.2} & \textbf{63.8} & \textbf{80.0} & \textbf{85.2} & \textbf{16.2} & \textbf{8.2} & \textbf{25.3} & \textbf{33.4} & \textbf{47.0} & \textbf{59.5} & \textbf{77.1} & \textbf{83.2} \\ 
\toprule
\toprule
FrameNet~\cite{Huang19FrameNet} & 14.7  & 7.7    & 22.8  & 33.4     & 48.9    & 62.5    & 80.1 & 85.8 & 18.6  & 11.0   & 26.8  & 23.2     & 36.6    & 50.7    & 72.0 & 79.5 \\
VPLNet~\cite{wang2020vplnet} & 13.8  & 6.8    & -  & -     & -    & 66.3    & 81.8 & 87.0 & 18.0  & 9.8   & -  & - & -    & 54.3    & 73.8 & 80.7 \\
DFPN+TAL & \textbf{12.6}  & \textbf{6.0}    & \textbf{21.1}  & \textbf{42.8} & \textbf{57.5}  & \textbf{69.3}    & \textbf{83.9} & \textbf{88.6} & \textbf{16.1}  & \textbf{8.1}    & \textbf{25.1}  & \textbf{33.6}     & \textbf{47.3}    & \textbf{59.8}    & \textbf{77.4} & \textbf{83.4} \\ \bottomrule

\end{tabular}}
\vspace{4mm}
\caption{(Top) Comparison between different loss functions. (Bottom) Comparison with the state-of-the-art methods.}
\label{table:loss_statistics}
\end{table}

\section{Summary}
In this paper, we present a new spatial rectifier to estimate surface normals from a tilted image. The spatial rectifier is learned to warp the tilted image to the rectified image such that its surface normal distribution is matched to that of the training data. We train the spatial rectifier jointly with the surface normal estimator by synthesizing tilted images from the ScanNet dataset. To facilitate practical deployment of surface normal estimation, we design a new efficient network that produces the state-of-the-art accuracy while maintaining lower computational burden.
Further, we propose a truncated angular loss that  addresses the limitations of the $L_2$ loss, resulting in accurate estimation, especially in the region of small angular error. Our method outperforms the state-of-the-art baselines not only on ScanNet and NYUv2 but also on a new Tilt-RGBD dataset that includes large roll and pitch camera motions.

\noindent \textbf{Acknowledgements} This work is supported by NSF IIS-1328722 and NSF CAREER IIS-1846031.
\clearpage
%
%

\bibliographystyle{splncs04}
\bibliography{references}
\end{document}


\pagestyle{headings}
\mainmatter
\def\ECCVSubNumber{1886}  

\title{Supplementary Material: \\ Surface Normal Estimation of \textit{Tilted} Images\\via Spatial Rectifier} 

\titlerunning{ECCV-20 submission ID \ECCVSubNumber} 
\authorrunning{ECCV-20 submission ID \ECCVSubNumber} 
\author{Anonymous ECCV submission}
\institute{Paper ID \ECCVSubNumber}

\maketitle



In this supplementary material, we first provide more details for obtaining supervision on the gravity $\mathbf{g}$ and the rectified direction $\mathbf{e}^{*}$ in Section~\textcolor{red}{3.1}.
Next, we compare the gradient characteristics of $L_2$/cosine and our proposed truncated angular losses (TAL) in Section~\textcolor{red}{3.3}.
Subsequently, training detail and choice of hyperparameters is discussed.
Finally, we provide additional qualitative surface normal prediction on various scenes.

\section{Optimal Principle Direction}


As described inSection~\textcolor{red}{3.1}, we supervise the spatial rectifier with not only the ground-truth surface normal, but also the gravity $\mathbf{g}$ and rectified principle direction $\mathbf{e}$.
While the former can be extracted from the 3D mesh reconstruction (ScanNet) or IMU, the latter can be found by solving the following optimization problem:
\begin{align}
    \mathbf{e}^* = \underset{\mathbf{e}}{\operatorname{argmin}}~D_{KL}\left(P(\mathbf{e})||Q\right) + \lambda_{\mathbf{e}}V\left(\overline{\mathcal{I}}\left(W_{\mathbf{g},\mathbf{e}}\right)\right), \label{Eq:e}
\end{align}
where $Q$ is the surface normal's training data distribution, $P(\mathbf{e})$ is the transformed distribution of the rectified image $\overline{\mathcal{I}}$, with the warping operation $W_{\mathbf{g},\mathbf{e}} = \mathbf{K}\mathbf{R}\mathbf{K}^{-1}$ where $\mathbf{K}$ is the camera matrix, and $\mathbf{R}$ is the shortest rotation matrix that brings $\mathbf{g}$ to $\mathbf{e}$ as follows:
\begin{align}
     \mathbf{R} = \mathbf{I}_{3} + 2\mathbf{e}\mathbf{g}^\mathsf{T} 
     - \frac{1}{1 +\mathbf{e}^\mathsf{T}\mathbf{g}} \left(\mathbf{e} + \mathbf{g}\right)\left(\mathbf{e}+\mathbf{g}\right)^\mathsf{T}\label{eq:rot_matrix_gravity_alignment} 
\end{align}
In addition, the Kullback-Leibler (KL) divergence is computed as:
\begin{align}
    D_{KL}\left(P(\mathbf{e})||Q\right) = \sum_{\mathbf{n} \in \mathds{S}^2} P\left(\mathbf{R}\mathbf{n}\right) log\left(\frac{Q(\mathbf{n})}{P(\mathbf{R}\mathbf{n})}\right) \label{eq:kld} 
\end{align}
where $\mathbf{R}$ is parameterized with $\mathbf{e}$ as defined in Equation~\eqref{eq:rot_matrix_gravity_alignment} and the area of the rectified image $V\left(\overline{\mathcal{I}}\left(W_{\mathbf{g},\mathbf{e}}\right)\right)$ can be obtained from the 2D transformed coordinates of the top-left (A), top-right (B), bottom-right (C), bottom-left (D) image's corners as:
\begin{align}
    V\left(\overline{\mathcal{I}}\left(W_{\mathbf{g},\mathbf{e}}\right)\right) = \frac{1}{2}\left(\begin{vmatrix} x_A & x_B \\ y_A & y_B \end{vmatrix} + \begin{vmatrix} x_B & x_C \\ y_B & y_C \end{vmatrix} + \begin{vmatrix} x_C & x_D \\ y_C & y_D \end{vmatrix} + \begin{vmatrix} x_D & x_A \\ y_D & y_A \end{vmatrix} \right)\label{eq:area} 
\end{align}
where each 2D rectified coordinate of (A, B, C, D) is computed as:
\begin{align}
    \begin{bmatrix} x \\ y \end{bmatrix} = \Pi\left( \mathbf{K}\mathbf{R}\begin{bmatrix} u \\ v \\ 1\end{bmatrix}\right) \label{eq:coordinates} 
\end{align}
where $\begin{bmatrix} u \ v \end{bmatrix}^{T}$ is the homogeneous coordinate of each corner and $\Pi(.)$ is the pinhole projection operator.

We solve for the optimal rectified direction $\mathbf{e}^{*}$ by stochastic gradient descent algorithm using the closed-form gradient.
First, for the KL-divergence, we parameterize the image's surface normal distribution $P$ as Gaussian mixture model with $k$ modes, then the gradient can be computed as:
\begin{align}
    \nabla_{\mathbf{e}} D_{KL}\left(P(\mathbf{e})||Q\right) = \sum_{\mathbf{n} \in \mathds{S}^2} \nabla_{\mathbf{e}} P\left(\mathbf{Rn}\right) \left(log \ P\left(\mathbf{Rn}\right) - log \ Q(\mathbf{n})\right)
\end{align}
where $\nabla_{\mathbf{e}} P\left(\mathbf{Rn}\right)$ can be computed by chain-rule:
\begin{align}
    \nabla_{\mathbf{e}} P\left(\mathbf{Rn}\right) &= \nabla_{\mathbf{e}} (\mathbf{Rn}) \ \nabla_{\mathbf{x}} \left(P\left( \mathbf{x} \right)\right) |_{\mathbf{x} = \mathbf{Rn}}, \\
    \nabla_{\mathbf{e}} \mathbf{Rn} &= \left( 2\mathbf{g}^{T}\mathbf{n} - \frac{\mathbf{n}^{T}(\mathbf{e}+\mathbf{g})}{1+\mathbf{e}^{T}\mathbf{g}}\right)\mathbf{I} \nonumber \\ & \ \ \ \ - \frac{1}{1+\mathbf{e}^{T}\mathbf{g}}\mathbf{n}(\mathbf{e}+\mathbf{g})^{T} + \frac{\mathbf{n}^{T}(\mathbf{e}+\mathbf{g})}{(1+\mathbf{e}^{T}\mathbf{g})^{2}}\mathbf{g}(\mathbf{e}+\mathbf{g})^{T} \label{eq:rot_grad}
\end{align}
Finally, we obtain the gradient of $V$ as follows:
\begin{align}
    \nabla_{\mathbf{e}} V = \mathbf{K}^{T} \sum_{i \in \{A, B, C, D\}} \nabla_{\mathbf{e}} \left( \mathbf{R}\begin{bmatrix} u_i \\ v_i \\ 1\end{bmatrix} \right) \nabla_{x_i, y_i} V(x_i, y_i)
\end{align}
where the gradient of $\mathbf{e}$ involved rotation matrix is computed similarly to Equation~\eqref{eq:rot_grad}\footnote{We project the gradient into the orthogonal direction of $\mathbf{e}$ and normalizing $\mathbf{e}$ at every iteration.} and the gradient of $V$ w.r.t each 2D coordinate $x_i, y_i$ can be computed in a straightforwardly.

\section{Gradient comparison $L_2$ and truncated angular loss}

We recall the $L_2$ loss described in Section~\textcolor{red}{3.3} as:
\begin{align}
 \mathcal{L}_{L_2} &= \sum_{\mathcal{D}}
 \sum_{\mathbf{x} \in \mathcal{R}(\mathcal{I})} \| \mathbf{n}_{\mathbf{x};\mathcal{I}}- \widehat{\mathbf{n}}_{\mathbf{x};\mathcal{I}} \|_{2}^{2} = \sum_{\mathcal{D}}
 \sum_{\mathbf{x} \in \mathcal{R}(\mathcal{I})} 1 - \mathbf{n}_{\mathbf{x};\mathcal{I}}^{\mathsf{T}}\widehat{\mathbf{n}}_{\mathbf{x};\mathcal{I}} \label{eq:traditional_supervised_loss}
\end{align}
where $\mathbf{n}_{\mathbf{x};\mathcal{I}} = f(\mathbf{x}; \mathcal{I})$ is the estimated surface normal, $\widehat{\mathbf{n}}_{\mathbf{x};\mathcal{I}}$ is its ground truth.
The gradient of $L_2$ loss is the sum of each surface normal prediction term:
\begin{align}
    \nabla_{\mathbf{n}} \left( 1 - \mathbf{n}^{T}\widehat{\mathbf{n}}\right) = \left( \mathbf{I} - \mathbf{n}\mathbf{n}^{T} \right) \widehat{\mathbf{n}}
\end{align}
where we can observe that the gradient magnitude converges to $0$ whenever $e_{\mathbf{x};\mathcal{I}} = \mathbf{n}^{T}\widehat{\mathbf{n}}$ is close to $1$.
In contrast, the angular loss experiences an exploding gradient at the same region:
\begin{align}
    \nabla_{\mathbf{n}} cos^{-1}\left(\mathbf{n}^{T}\widehat{\mathbf{n}}\right) = \frac{-1}{\sqrt{(1-\mathbf{n}^{T} \widehat{\mathbf{n}})(1+\mathbf{n}^{T} \widehat{\mathbf{n}})}}\left( \mathbf{I} - \mathbf{n}\mathbf{n}^{T} \right) \widehat{\mathbf{n}}
\end{align}
We found that this loss's gradient characterisitic is suitable for training the surface normal prediction network on noisy and outlier contaminated data.
Moreover, we prevent the same exploding gradient effect at $e_{\mathbf{x};\mathcal{I}} \simeq -1$, since most of the ground-truth data around this area are outlier, by employing the $L_2$ loss.
Therefore, the final proposed truncated angular loss enjoys the same gradient characteristic at $e_{\mathbf{x};\mathcal{I}} \simeq 1$ and is robust to outlier data at $e_{\mathbf{x};\mathcal{I}} \simeq -1$ (see Section~\textcolor{red}{3.3}).

\section{Spatial Rectifier training details and qualitative results}

Due to its 
\begin{itemize}
    \item Interesting warping directions of our method even if the gravity is incorrectly predicted.
\end{itemize}
\clearpage
%
%

\bibliographystyle{splncs04}
\bibliography{references}